# Text Similarity From Image Contents using Statistical and Semantic Analysis Techniques

Sagar Kulkarni, Sharvari Govilkar and Dhiraj Amin

Department of Computer Engineering, Pillai College of Engineering, New Panvel, India

*ABSTRACT*

*Plagiarism detection is one of the most researched areas among the Natural Language Processing(NLP) community. A good plagiarism detection covers all the NLP methods including semantics, named entities, paraphrases etc. and produces detailed plagiarism reports. Detection of Cross Lingual Plagiarism requires deep knowledge of various advanced methods and algorithms to perform effective text similarity checking. Nowadays the plagiarists are also advancing themselves from hiding the identity from being catch in such offense. The plagiarists are bypassed from being detected with techniques like paraphrasing, synonym replacement, mismatching citations, translating one language to another. Image Content Plagiarism Detection (ICPD) has gained importance, utilizing advanced image content processing to identify instances of plagiarism to ensure the integrity of image content. The issue of plagiarism extends beyond textual content, as images such as figures, graphs, and tables also have the potential to be plagiarized. However, image content plagiarism detection remains an unaddressed challenge. Therefore, there is a critical need to develop methods and systems for detecting plagiarism in image content. In this paper, the system has been implemented to detect plagiarism form contents of Images such as Figures, Graphs, Tables etc. Along with statistical algorithms such as Jaccard and Cosine, introducing semantic algorithms such as LSA, BERT, WordNet outperformed in detecting efficient and accurate plagiarism.*

## 1. INTRODUCTION

Plagiarism is a big problem in academics, researches and it can be a big problem in every department of the education sector. Plagiarism is defined as copying someone's work and presenting it as one's own work. The problem of plagiarism has become an important issue in the field of education and technology due to the wide use and availability of electronic devices and internet makes it easy for students, authors and even academic people to access and use any piece of information and embed it into his/her own work without proper citation. The objectives and importance of plagiarism detection are multifaceted. Firstly, plagiarism detection acts as a deterrent, discouraging individuals from engaging in plagiarism by creating awareness about the consequences and potential repercussions. By fostering a culture of originality and integrity, it upholds the values of academic and professional communities. Secondly, plagiarism detection serves as a proactive mechanism to identify and address instances of plagiarism before they tarnish the reputation of individuals or institutions. By promptly identifying and addressing plagiarism cases, it helps maintain the credibility and trustworthiness of scholarly and creative works.

Additionally, plagiarism detection supports the development of a scholarly and creative ecosystem based on trust, accountability, and respect for intellectual contributions. It encourages proper citation and attribution practices, thereby giving due credit to the original authors and creators. Furthermore, plagiarism detection ensures fair competition among researchers, writers, and creators, as it prevents the undue advantage gained by those who engage in plagiarism. Not only text but images such as Figures, Graphs, Tables, flowcharts etc. can be plagiarized. If the author has not mentioned the credit for the original author from where he/she copied the image then it is said to be plagiarized.

## 2. LITERATURE REVIEW

Most of the research has been carried out with statistical approaches using bag of word and dictionary-based approaches for detecting similarity between the text. Plagiarism Detection for textual contents with semantic algorithms. Already we have understood and implemented statistical based approaches for textual similarity identification and we concluded that the use of only type statistical approaches will not be fair enough. It is required to introduce algorithms which will check the textual contents and find semantic relation to the actual contents of other documents to determine accurate plagiarism. A part of Image plagiarism detection has already been implemented and presented in a previous seminar. The prior system was able to detect Plagiarism of image when it is compared with another single image. The actual system will perform if we compare the suspicious image contents with the images stored in the database as a novel. In this a single suspicious image will be compared with all of the images and it is a convenient and accurate way to check whether the image is plagiarized or not.

The proposed system in [1] classifies the image and then extracts the text data from them. Image Classification is done using the Convolutional Neural Network, which classifies images based on feature extraction. Text extraction is done using the Python-Tesseract OCR package. The text extraction from images is very important to our research also so as to detect plagiarism in images too. The authors in [2] have developed a text extraction model for image documents, using a combination of the two powerful methods Connected Component and Edge Based Method, in order to enhance performance and accuracy of text extraction authors have used an integrated simulation tool.

There are many research papers available which perform textual plagiarism detection. But checking plagiarism in only text is not sufficient. There are also a few other areas (such as Tables, Figures, Graphs, Images, Citations etc.) where plagiarism can happen. Through our rigorous literature survey, we haven't found any such kind of research which detects plagiarism from non-textual data too. In one of our own previous papers [3], we have attempted to design the Image to text conversion through API and detecting the text similarity with statistical methods. Over the period of time in the last six months we realized that the API that we were using was capable of accepting images which are stored on cloud/ server. But for our research it is very important that the system must accept the input image which is stored in the computer memory disk. Also, an important point to consider is that the system must be trained enough with a large number of images which can further be capable of understanding the suspicious image if any which is plagiarized.

The authors in [4] have made an attempt to detect the text from the images based on Natural scene. The authors agree that there is no single system which will accept images containing different languages and the system will interpret it and extract text. So, it is very important while designing any method to consider the language or script of the text which is embedded in the images. There is no perfect one stop solution for recognizing the script of all the languages using a single method. The deep learning and machine learning approaches can be useful for achieving this task.Another approach of using Fuzzy rule-based decision and artificial neural network to detect the characters and numbers has presented in [5]. The system is designed which works for detecting characters and numbers present in the images by using pure image processing techniques the system is designed to detect the handwritten textual part of the images.

In [6], the authors have proved that Microsoft vision took much less time to recognize the text from the images. Microsoft was 37% faster than Google and 3.15 times faster than Amazon. Microsoft had no language filter on its text extraction. It kept incorrectly identifying words and characters from non-English languages, which severely hurt its accuracy score. Again, the text extraction needs a lot more manual evaluation and tweaking to ensure that results are scored accurately. In our research we are planning to detect image plagiarism where images contain textual information in English script. Google is good for accuracy, Amazon is good for cost,and Microsoft is good for speed. For text processing all three do not have big differences so we selected to go with Microsoft API as it's quite faster than the other two. Also, there is a facility with Microsoft vision to use API free of cost for 5000 attempts/month which is not the case with others.

The demonstration of use of Google cloud API to extract text from images is given in [7]. It has been observed that google computer vision is providing a good platform to detect the text contained from the image. The accuracy of the system is also acceptable. When we tried to adopt the model in our project then we came to know that the Google computer vision API requires you to subscribe which credit card details and also after a few free trials, it will be charged automatically. Thus, we decided that the same results we can obtain from Microsoft Vision API as well like google computer vision API. So, we configured the system with Microsoft Vision API and observed the accuracy of the system. The Microsoft Vision API is also performing best when the images are in proper readable format. If the image is skewed and having blur text then Microsoft Vision API gives improper output.

The intention of this literature review was to look for the best suitable approach to perform Image to text conversion and detecting plagiarism within the images if any. Overall, there are many image processing approaches that can extract text from the images such as Edge based, Component based etc. But doing so was dragging our main objective of the research towards the Image processing part. Thus, we were looking for an easy and better solution to get textual contents from the images. For this we came across the API usage by which we can achieve the satisfactorily accurate text contents from the images. After the literature survey we decided to use Microsoft vision API to extract textual information from the images.

The authors in [8], proposed a system that classifies the image and then extracts the text data from them. Image Classification is done using the Convolutional Neural Network, which classifies images based on feature extraction. Text extraction is done using the Python-Tesseract OCR package. The user interface is required to be designed so that users can upload the image files into the system.

Another approach of extracting text from images is presented in [9]. Text extraction from image documents has been done using a combination of the two powerful methods Connected Component and Edge Based Method. Finally, the extracted and recognized words are converted to speech for proper use for visually impaired people. The system is more of a pure image processing system. In [10], it has been observed and proved that with Google cloud vision API the output is quite satisfactory when the images are clear and without any noise but when the noise is added into the images the Google Cloud API is unable to detect images or even the text within those images. The API is unable to recognize the text when the noise level is from 10 % and above. Thus, the selection of Microsoft API is still worth adopting in our research

## 3. SYSTEM OVERVIEW

There are high chances that the plagiarist will take advantage of the loopholes and copy existing images/drawings/graphs as it is in their own research articles. Image plagiarism is a serious issue and needs to be addressed as it is important as the text plagiarism detection. The experimentation will involve a diverse dataset containing various image formats like jpg, jpeg, png, bmp for image content plagiarism detection.

**Conversion of Image to Text**
the system will extract the textual contents from the images using Microsoft API. The extracted text will be used for similarity identification between the text contents of other images in the corpus. The intention is to check the contents of those images or tables with the contents of the suspicious images or tables and not the shape of the images or tables.

**Input Documents**
A suspicious English text document will be given as input to the system. The system will compare the similarity of this document with all the reference documents and produce a plagiarism report.

**Pre-processing and NLP Operations**
The system is extended and implemented in such a way that it will read all the corpus images and extract all the textual data from images which are stored in the separate files.

**Text Similarity Identification**
When a suspicious image is given as an input to the system then the system will extract the text contents from it and with the help of a few advanced algorithms it will compare the text with the corpus contents.

**Percentage of Plagiarism Contents**
The percentage of similarity depends on the amount of text contents copied in suspicious images from the original corpus images.

**Plagiarism detection from Images/Graphs/Tables**
Plagiarism detection within image contents or drawings or tables is an important area that needs to be considered as seriously. There is a high possibility that the plagiarist copies the images/drawings from existing articles by changing the orientation of images. Thus, it is required to check plagiarism from non-textual things from the articles. Our system is upgraded to check plagiarism by comparing suspicious images with all corpus images at single time. This image plagiarism system is designed with two methods: cosine similarity and vectored TF-IDF similarity method. The system creates the vocabulary of all the words present in the actual contents of all the images. This vocabulary is used to compare the textual content of suspicious images with the corpus collection.The vectored TF-IDF method creates a vector of each image content which is extracted and saved as a separate document in the Corpus. When a suspicious image is given as an input then the Microsoft Vision will be used to extract the textual content from that image and for the textual contents, TF-IDF is created which will be further used for comparing with all the vectors of corpus collection documents.

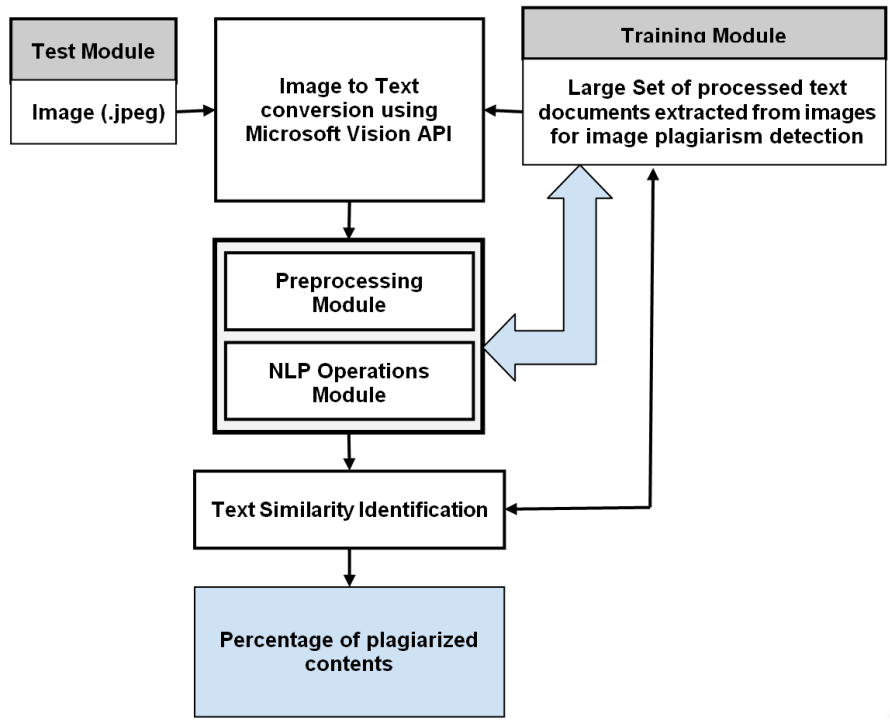

Fig 1: Plagiarism detection for Images and Tables.

## 4. IMPLEMENTATION

It is likely that when we compare one to one image then plagiarism may not be found but if we compare the suspicious image with all images in the corpus collection then the results can be notably different. Also the plagiarism detection within table contents and charts/graphs has been checked as it has been asked by subject matter experts. This image content plagiarism detection system is designed with two methods: cosine similarity and vectored TF-IDF similarity method. The system creates the vocabulary of all the words present in the actual contents of all the images. This vocabulary is used to compare the textual content of suspicious images with the corpus collection.

The vectored TF-IDF method creates a vector of each image content which is extracted and saved as a separate document in the Corpus. When a suspicious image is given as an input then the Microsoft Vision will be used to extract the textual content from that image and for the textual contents, TF-IDF is created which will be further used for comparing with all the vectors of corpus collection documents. The sample output of plagiarism percentage found with an image using vectored TF-IDF is as shown in Figure 3.

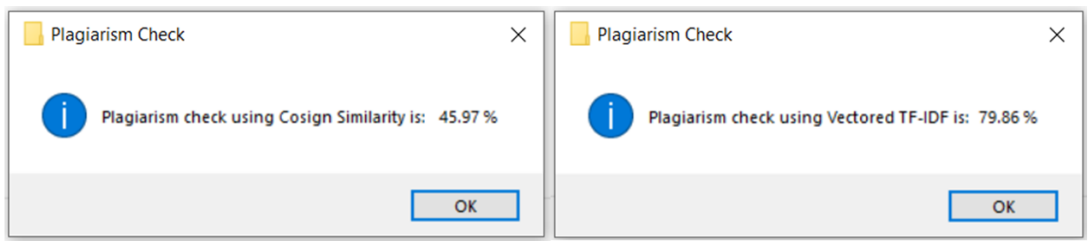

Fig 2. Plagiarism % using Cosine method    Fig 3. Plagiarism % using Vectored TF-IDF method

Table 1 represents all such scores that we had tested against various images. The plagiarism percentage using Cosine and Vectored TF-IDF are listed in the given table.

Table 1. Plagiarism % of sample test images/drawings

| Input | Plagiarism % using Cosine Method | Plagiarism % using Vectored TF-IDF Method |
|---|---|---|
| Drawing 1 | 41.59 | 72.00 |
| Drawing 2 | 32.40 | 54.57 |
| Drawing 3 | 30.84 | 49.71 |
| Drawing 4 | 32.40 | 54.57 |
| Table 1 | 45.97 | 79.86 |
| Table 2 | 45.12 | 79.14 |
| Table 3 | 40.46 | 69.57 |
| Table 4 | 36.36 | 65.57 |
| Graph 1 | 42.22 | 75.29 |
| Graph 2 | 41.87 | 74.00 |
| Graph 3 | 40.60 | 73.43 |
| Graph 4 | 40.74 | 73.86 |

Table 2 shows the effect of plagiarism results when a suspicious image is a totally novel image. The input image is not plagiarized at all and still we are checking it against all the images in the corpus then the same methods can show the less percentage as there is as good as no plagiarized contents in the input image.

Table 2. Result analysis of Plagiarism check for Dis-similar Images/Graphs/Tables

| Input | Plagiarism % using Cosine Method | Plagiarism % using Vectored TF-IDF Method |
|---|---|---|
| Drawing 1 | 1.92 | 0.67 |
| Drawing 2 | 1.62 | 2.27 |
| Table 1 | 3.76 | 1.32 |
| Table 2 | 7.39 | 1.48 |
| Graph 1 | 7.52 | 1.50 |
| Graph 2 | 5.44 | 1.09 |

After implementing the two methods such as Cosine and Vectored TF-IDF, the system is enhanced with introducing semantic algorithms such as LSA, BERT, and WordNet etc. to obtain more accurate results in plagiarism detection. To perform semantic analysis the system has to preprocess the input image file. Preprocessing steps, including tokenization, stop word analysis, lemmatization, NER, and reference removal, are applied. When given an input image file, the system compares the extracted text with trained image contents to detect plagiarism. Table 3 shows the plagiarism percentages for sample image files.

Table 3. Image Content Plagiarism Detection: Analysis of Sample Input

| Sr.No. | Algorithm | Plagiarism % when Named Entities Included | Plagiarism % when Named Entities Excluded |
|---|---|---|---|
| 1 | Jaccard | 31.99 | 25.24 |
| 2 | Cosine | 17.35 | 15.34 |
| 3 | LSA | 27.84 | 18.26 |
| 4 | BERT | 18.81 | 18.44 |
| 5 | WordNet | 14.82 | 18.8 |

Named Entity Recognition (NER) is a vital technique in NLP with applications in information retrieval and plagiarism detection. NER identifies and classifies named entities like person names, organization names, locations, and dates in text data. In plagiarism detection, NER is particularly useful for detecting disguised plagiarism, where content is modified. By leveraging NER, researchers can accurately identify disguised plagiarism by comparing named entities across documents.

In the analysis of the sample input image, it becomes evident that excluding named entities provides satisfactory results for plagiarism detection. Despite the possibility of a slightly higher percentage of detected plagiarism when excluding named entities, it is still advisable to adopt this approach due to its effectiveness in removing potential false positives. The sample results using all the algorithms utilized, providing the plagiarism percentage with and without named entity recognition (NER) inclusion. This allows for a comprehensive understanding of the extent of plagiarism detected in figures, tables, and graphs. The system has been tested with multiple image files such as figures, flowcharts, graphs, tables etc. to check the efficiency of the algorithms in detecting the accurate plagiarism. This plagiarism check is performed after removing named entities as it is already provided that exclusion of named entities gives more accurate and efficient results in plagiarism detection. Table 4 allows for a comparison of the percentage of plagiarism of figure/flowcharts images given by algorithms when named entities are excluded.

Table 4. Plagiarism Percentage with Exclusion of Named Entities for Sample Input Figures

| Input | Jaccard | Cosine | LSA | BERT | WordNet |
|---|---|---|---|---|---|
| Figure1 | 33.07 | 18.2 | 49.89 | 22.47 | 25.83 |
| Figure2 | 27.24 | 16.52 | 27.98 | 24.79 | 21.04 |
| Figure3 | 21.36 | 14.02 | 37.46 | 19.91 | 22.96 |
| Figure4 | 36.89 | 18.76 | 23.42 | 21.13 | 26.2 |
| Figure5 | 25.24 | 15.34 | 18.26 | 18.44 | 18.8 |

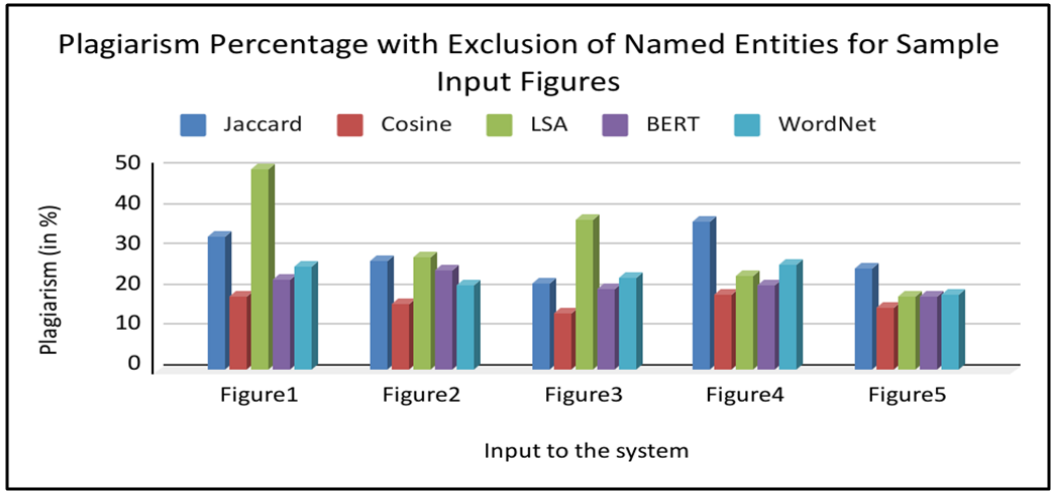

Fig. 4. Visualizing Plagiarism Percentage for Sample Input Figures with Named Entities Excluded

Table 5 shows the plagiarism percentage obtained for other image types such as Graphs and Tables. By observing the outcomes of each algorithm, it can be clearly seen that semantic algorithms outperforms in detecting accurate plagiarism.

Table 5. Plagiarism Percentage with Inclusion of Named Entities for Sample Input Tables

| Input  | Jaccard | Cosine | LSA   | BERT  | WordNet |
|--------|---------|--------|-------|-------|---------|
| Table1 | 77.67   | 27.58  | 25.2  | 29.9  | 17.23   |
| Table2 | 42.8    | 20.71  | 12.23 | 26.47 | 20.67   |
| Table3 | 46.6    | 21.19  | 13.69 | 27.68 | 25.86   |
| Table4 | 11.07   | 33.04  | 17.24 | 29.94 | 21.04   |
| Table5 | 44.66   | 20.73  | 13.72 | 30.51 | 25.86   |
| Graph1 | 27.18   | 15.96  | 27.4  | 40.91 | 17.23   |
| Graph2 | 23.35   | 15.29  | 27.11 | 40.74 | 17.23   |
| Graph3 | 23.35   | 15.29  | 13.39 | 33.83 | 17.23   |
| Graph4 | 33.07   | 18.2   | 15.81 | 31.18 | 17.23   |
| Graph5 | 28.36   | 12.09  | 18.65 | 27.21 | 25.86   |

The overall system shows the plagiarism detection from images such as figures, graphs, tables etc. At all times it has been observed that by introducing semantic algorithms, the system outperforms in detecting the accurate plagiarized contents from the images, drawings and tables if any. LSA and BERT algorithms have shown superior performance in graph plagiarism detection, especially when named entities are excluded. Despite a slightly higher plagiarism percentage in some cases, excluding named entities helps reduce false positives.

## 5. CONCLUSION

This research covers image content plagiarism detection in which multiple algorithms such as Vectored TF-IDF, Jaccard, Cosine, LSA, BERT, WordNet, etc., were implemented to compare plagiarism detection results. The emphasis was on semantic analysis and resolving named entities. The system works with diverse image formats such as .jpg, .png, .bmp etc. as part of the input files for image content plagiarism detection. The preprocessing of documents and images is crucial to reduce the comparison time during the plagiarism detection process. The overall findings of the research indicate that semantic methods especially LSA and BERT outperform other methods in terms of efficiency and accuracy for plagiarism detection as compared with statistical algorithms such as Jaccard, Cosine, Vectored TF-IDF etc. By leveraging semantic analysis, handling named entities and resolving references the system achieved more efficient and accurate detection of plagiarized content across languages.

## REFERENCES

[1] Deepa, R., & Lalwani, K. N. (2019), "Image Classification and Text Extraction using Machine Learning", 2019 3rd International Conference on Electronics, Communication and Aerospace Technology (ICECA). doi:10.1109/iceca.2019.8821936,

[2] K.N. Natei, J. Viradiya, S. Sasikumar. "Extracting Text from Image Document and Displaying Its Related Information", K.N. Natei Journal of Engineering Research and Application, ISSN : 2248-9622, Vol. 8, Issue5 (Part -V) May 2018, pp 27-33

[3] Rohini Pimparkar, Nehal Shinde, Vivek Thakur, Prof. Sagar Kulkarni, "IDENTIFICATION OF TEXTUAL SIMILARITY FROM IMAGE CONTENTS USING SEMANTIC ANALYSIS", International Research Journal of Modernization in Engineering Technology and Science Volume:03/Issue:04/April-2021

[4] Shiravale S. S. Sannakki S. S. and Rajpurohit V. S., "Recent Advancements in Text Detection Methods from Natural Scene Images", International Journal of Engineering Research and Technology. ISSN 0974-3154, Volume 13, Number 6 (2020), pp. 1344-1352

[5] V. Lakshman Narayana, B. Naga Sudheer, Venkata Rao Maddumala, P.Anusha, "FUZZY BASE ARTIFICIAL NEURAL NETWORK MODEL FOR TEXT EXTRACTION FROM IMAGES", Journal of Critical Reviews, ISSN- 2394-5125 Vol 7, Issue 6, 2020

[6] Jake Singh, Jackson Wheeler et. al. "A Comparison of Public Cloud Computer Vision Services", OSF Storageosf.io/9t5qf, United states, Formal Report, 2019

[7] CARLOS D'ANDRÉA, ANDRÉ MINTZ. "Studying the Live Cross-Platform Circulation of Images With Computer Vision API: An Experiment Based on a Sports Media Event", International Journal of Communication 13(2019), 1825–1845, 1932–8036/20190005

[8] R. Deepa, Kiran N Lalwani, "Image Classification and Text Extraction using Machine Learning", Proceedings of the Third International Conference on Electronics Communication and Aerospace Technology [ICECA 2019] IEEE Conference Record # 45616; IEEE Xplore ISBN: 978-1-7281-0167-5

[9] K.N. Natei, J. Viradiya, S. Sasikumar, "Extracting Text from Image Document and Displaying Its Related Information", K.N. Natei Journal of Engineering Research and Application www.ijera.com ISSN: 2248-9622, Vol. 8, Issue5 (Part -V) May 2018, pp 27-33.

[10] Hossein Hosseini, Baicen Xiao and Radha Poovendran, "Google's Cloud Vision API Is Not Robust To Noise", 2017 16th IEEE International Conference on Machine Learning and Applications, DOI 10.1109/ICMLA.2017.0-172, 2017

[11] Nagoudi, E. M. B., Cherroun, H., &Alshehri, A. (2018). Disguised plagiarism detection in Arabic text documents. 2018 2nd International Conference on Natural Language and Speech Processing (ICNLSP). doi:10.1109/icnlsp.2018.8374395

[12] Vrbanec, T., & Mestrovic, A. (2017). The struggle with academic plagiarism: Approaches based on semantic similarity. 2017 40th International Convention on Information and Communication Technology, Electronics and Microelectronics (MIPRO). doi:10.23919/mipro.2017.7973544

[13] Schneider, johannes, Shepherd, D., Damevski, K., vomBrocke, J., & Bernstein, A. (2017). Detecting Plagiarism based on the Creation Process. IEEE Transactions on Learning Technologies, 1–1. doi:10.1109/tlt.2017.2720171

[14] Suleiman, D., Awajan, A., & Al-Madi, N. (2017). Deep Learning Based Technique for Plagiarism Detection in Arabic Texts. 2017 International Conference on New Trends in Computing Sciences (ICTCS). doi:10.1109/ictcs.2017.42

[15] Hanane, E., Erritali, M., &Oukessou, M. (2016). Semantic Similarity/Relatedness for Cross Language Plagiarism Detection. 2016 13th International Conference on Computer Graphics, Imaging and Visualization (CGiV). doi:10.1109/cgiv.2016.78

46   Computer Science & Information Technology (CS & IT)

## AUTHORS

**Sagar Kulkarni** has worked as assistant professor in the Computer Engineering department at Pillai College of Engineering, Mumbai, Maharashtra India. Currently he is pursuing Ph.D in Computer Engineering from University of Mumbai. He has completed M.E. in Computer Engineering from University of Mumbai, India. Sagar has received a BE degree in CSE from Shivaji University, Kolhapur. He has 15 years of experience in teaching. His areas of interest are text mining and summarization, system programming and compiler construction, natural language processing, and information retrieval. He can be contacted at email: sagark@student.mes.ac.in

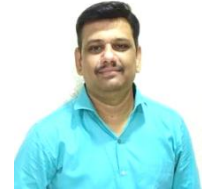

**Sharvari Govilkar** holds a PhD in information technology from University of Mumbai, India. She is a professor of Computer Engineering department at Pillai College of Engineering, Mumbai, Maharashtra, India. She has more than 26 years of teaching experience in the field of Computer Engineering. She is guiding many UG, PG projects and research scholars. She has published about 85 research papers in various national and international conferences and journals. She is also contributing as a reviewer for international conferences, journals, and transactions in the domain of artificial intelligence and natural language processing. Her areas of expertise are data science, natural language processing and social media text analytics. She can be contacted at email: sgovilkar@mes.ac.in.

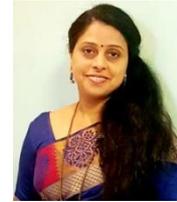

**Dhiraj Amin** received the ME degree in Computer Engineering from University of Mumbai, India, in 2015, respectively. He is currently working as assistant professor in the Computer Engineering department at Pillai College of Engineering, Mumbai, Maharashtra India. He has published several research papers in reputed international journals and conferences. He has more than 8 years of teaching experience and his areas of research include natural language processing, information retrieval, augmented reality, social media analytics and artificial intelligence. He can be contacted at email: amindhiraj@mes.ac.in.

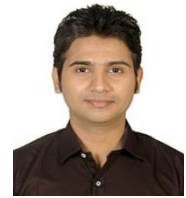